\documentclass[letterpaper, 10 pt, conference]{ieeeconf}  
\usepackage{censor}
\StopCensoring 

\IEEEoverridecommandlockouts                              

\makeatletter
\let\NAT@parse\undefined
\makeatother

\usepackage{graphics} 
\usepackage{amsmath} 
\usepackage{amssymb}  
\usepackage[hyphens]{url}
\usepackage{hyperref}
\usepackage[hyphenbreaks]{breakurl}
\usepackage{multirow}
\usepackage{tikz}

\newcommand\submittedtext{%
	\scriptsize This work has been submitted to the IEEE for possible publication. Copyright may be transferred without notice, after which this version may no longer be accessible.}

\newcommand\submittednotice{%

		\begin{tikzpicture}[remember picture,overlay]
			\node[anchor=south,yshift=0pt] at (current page.south) 	{\fbox{\parbox{\dimexpr0.95\textwidth-\fboxsep-\fboxrule\relax}{\submittedtext}}};
		\end{tikzpicture}%
}

\graphicspath{ {images/} } 

\title{\LARGE \bf From Transportation to Manipulation: \\ Enabling Grasping in Magnetic Robotics}

\author{\censor{Lara Bergmann$^{1}$, Noah Greis$^{1}$, Cedric Grothues$^{1}$, Lisa-Marie Weigelt$^{1}$, and Klaus Neumann$^{1,2}$}
\xblackout{\thanks{*This work was not supported by any organization}
\thanks{$^{1}$Lara Bergmann, Noah Greis, Cedric Grothues, Lisa-Marie Weigelt, and Klaus Neumann are with the Faculty of Technology, CITEC, Bielefeld University, Germany. Email corresponding author: {\tt\small {lara.bergmann@uni-bielefeld.de}}}%
\thanks{$^{2}$Klaus Neumann is with Fraunhofer IOSB-INA, Lemgo, Germany}}%
}

\begin{document}
\maketitle
\thispagestyle{empty}
\pagestyle{empty}
\submittednotice
\begin{abstract}
    Magnetic levitation (MagLev) systems have great potential for application in high-mix, low-volume manufacturing due to their scalability and flexibility, enabling highly reconfigurable in-machine material flow. However, their manipulation capabilities remain largely unexploited, as current applications almost exclusively focus on transportation. To enable grasping and manipulation directly on MagLev systems without requiring additional costly handling equipment, such as industrial robot arms, we present the \emph{Gripper MagBot}, a low-cost parallel 6-DoF manipulator with an integrated 1-DoF gripper that mechanically couples three MagLev movers. The \emph{Gripper MagBot} supports two operating configurations: a default mode and a single-track mode, selectable depending on the required stability and workspace footprint. To reconfigure a machine, the MagBot can be autonomously dropped off and picked up using a docking station. We showcase pick-and-place examples in simulation, as well as with the real \emph{Gripper MagBot} using our inverse kinematics controller. CAD files, assembly instructions, a component list, and videos are available at \url{https://sites.google.com/view/gripper-magbot}.
\end{abstract}
\section{INTRODUCTION}
\begin{figure}
	\centering
    \includegraphics[width=0.95\linewidth]{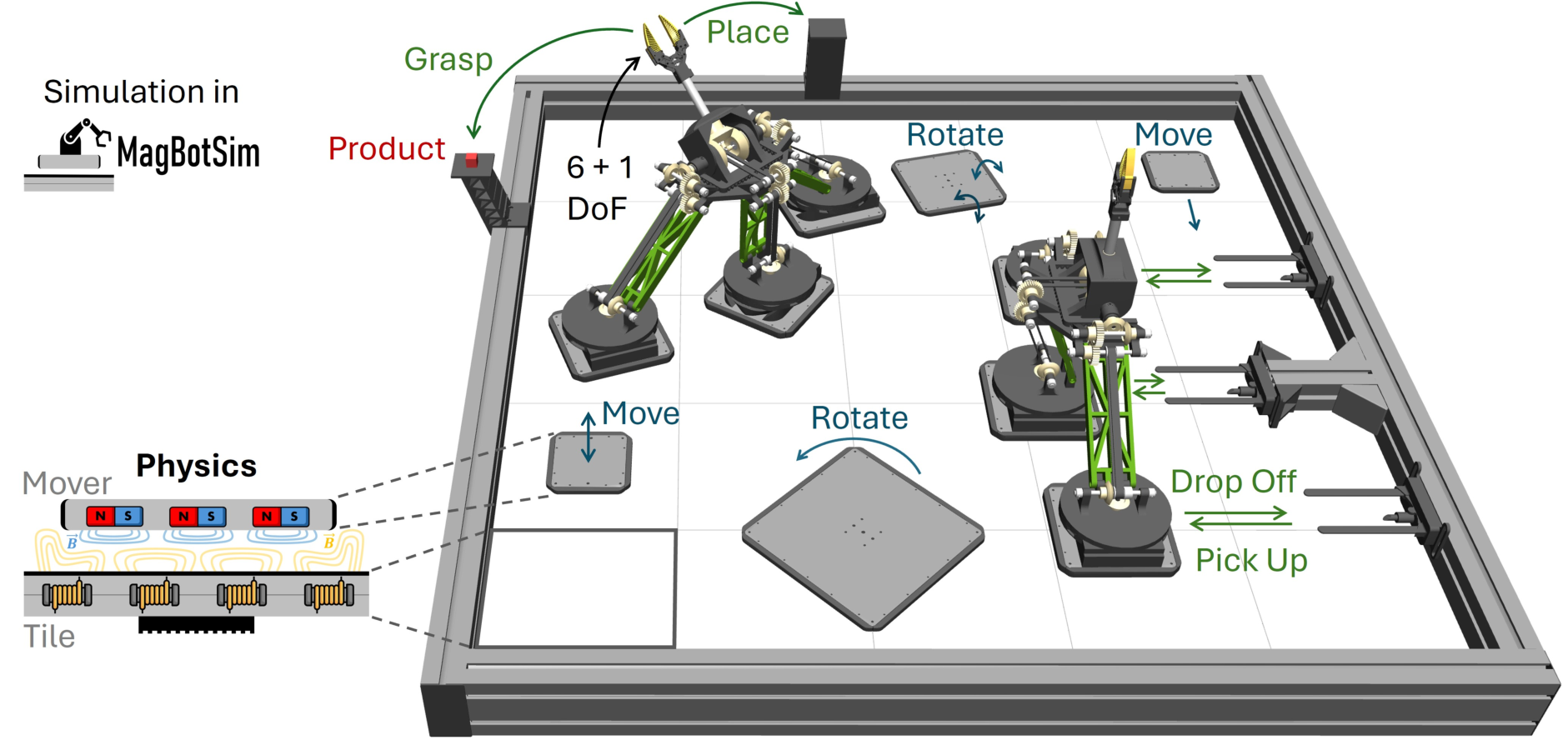}
        \caption{Schematic illustration of the \emph{Gripper MagBot} in the context of Magnetic Robotics. The movers can autonomously drop off or pick up the MagBot using a docking station.}
    \label{fig_visual_abstract}
\end{figure}

\begin{figure}
	\centering
    \includegraphics[width=0.9\linewidth]{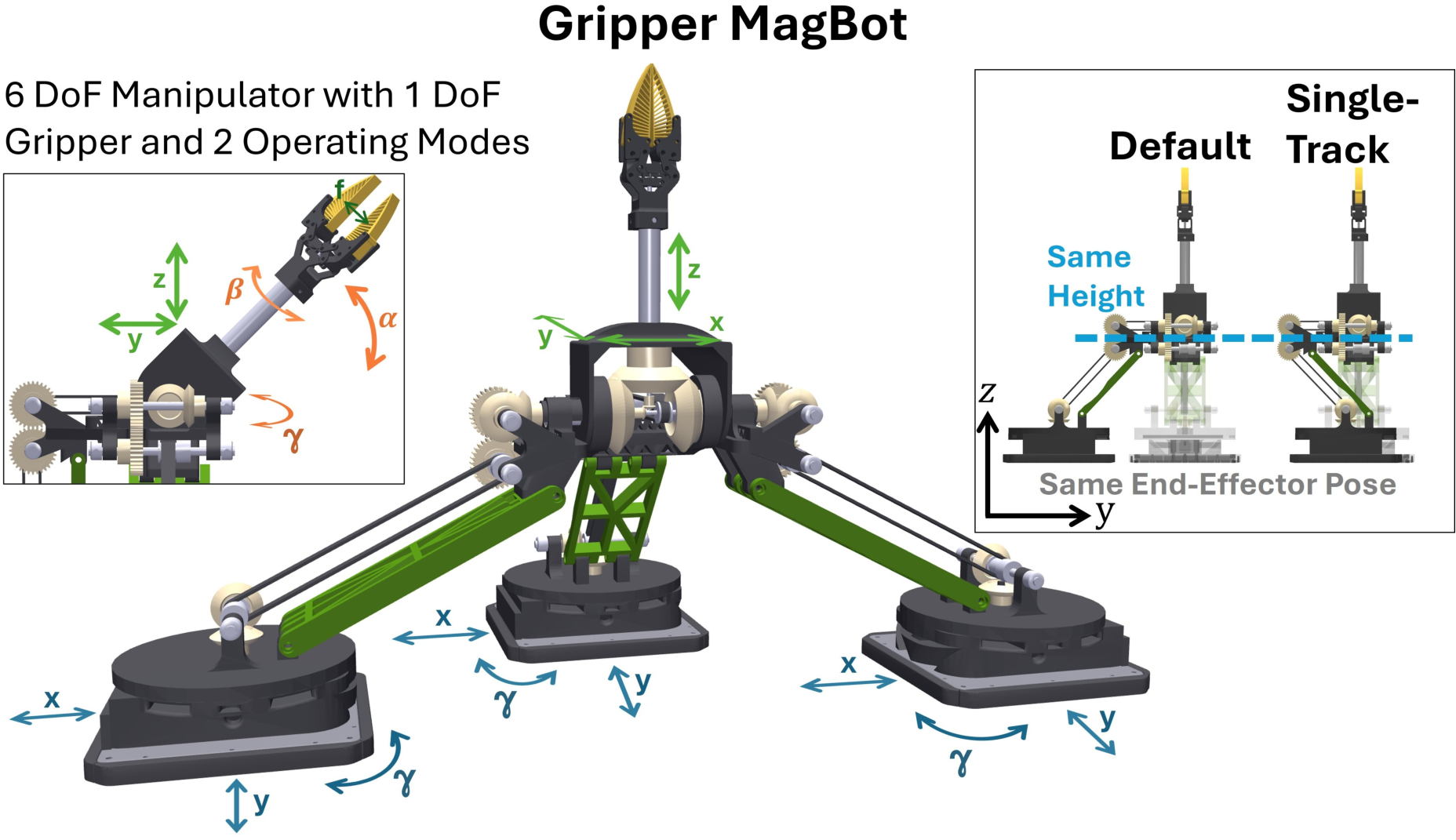}
        \caption{Our \emph{Gripper MagBot} has a 6-DoF manipulator and a 1-DoF gripper, which are mechanically coupled to the x-, y-, and $\gamma$-axes of the three movers. The MagBot has two operating modes, default and single-track, in which the same end-effector positions can be reached.}
    \label{fig_magbot_dofs_modes}
    \vspace{-0.3cm}
\end{figure}

\noindent
High-mix, low-volume production requires highly scalable and reconfigurable robotic systems that can be adapted to changing requirements. Magnetic Levitation (MagLev) systems inherently provide this adaptability, as these systems consist of a modularly configurable surface of static motor modules, referred to as \emph{tiles}, and actuated \emph{movers} of different sizes, each comprising permanent magnets arranged in Halbach arrays~\cite{lu_6d_2012}. All six degrees of freedom (DoFs) of the movers are controlled by electromagnetic fields generated by the tiles that interact with the Halbach arrays of the movers (see Fig.~\ref{fig_visual_abstract}). Commercial MagLev systems designed specifically for industrial applications include ACOPOS 6D (B\&R Industrial Automation), MagiFloater (RobustMotion), XBot (Planar Motor), ctrlX $\text{FLOW}^{\text{6D}}$ (Bosch Rexroth), and XPlanar (Beckhoff Automation). However, the manipulation capabilities of these systems remain largely unexploited, as a single mover is limited by its restricted workspace and lack of grasping ability. Therefore, in our previous work, we introduced the 6D-Platform MagBot~\cite{bergmann_transportation_2026}, a parallel kinematic mechanism with six DoFs that couples two movers, to increase the reachable workspace and payload compared with a single mover. However, the ability to grasp is still missing. To enable grasping based solely on the MagLev system without the need for additional, expensive handling equipment, such as robot arms, we present the \emph{Gripper MagBot} (see Fig.~\ref{fig_magbot_dofs_modes}), a low-cost parallel 6-DoF manipulator with a 1-DoF gripper that couples three MagLev movers.
Similar to the 6D-Platform MagBot, the \emph{Gripper MagBot} is compatible with all aforementioned MagLev systems and can be autonomously picked up or dropped off using the same docking mechanism. Therefore, future machines can intelligently reconfigure themselves by choosing the required manipulation capabilities and picking up the corresponding MagBot. In this paper, we make the following contributions:
\begin{itemize}
    \item Developed and built the \emph{Gripper MagBot}, which is compatible with the docking station of the 6D-Platform MagBot for autonomous reconfiguration
    \item Created an inverse kinematics controller
    \item Developed a physics-based simulation of the \emph{Gripper MagBot} integrated into MagBotSim~\cite{bergmann_magbotsim_2026}
    \item Investigated the gripper's performance metrics and evaluated its applicability
\end{itemize}
\section{RELATED WORK}
\noindent
\textbf{Magnetic Levitation.} 
MagLev is an active research field with very different directions. In the medical domain, magnetic microrobots are studied~\cite{isitman_trajectory_2025,sallam_autonomous_2024}, e.g. for autonomous object delivery~\cite{liu_on_2024}, which requires the manipulation of objects as well as proper trajectory planning. In contrast, our work focuses on MagLev systems for industrial automation, i.e. in a completely different field of application, which also requires a MagLev system on a larger scale. Another MagLev research direction investigates the low-level control of the movers, i.e. how to adjust the magnetic fields, and hardware development~\cite{hartmann_end--end_2025,lu_6d_2012}. Other works focus on the high-level control, e.g. trajectory planning, with industrial MagLev systems~\cite{pierer_von_esch_sensitivity-based_2025,janning_conflict-based_2025,tistaert_multi-agent_2026}, whereas this work investigates object manipulation, which is only rarely studied in the current literature. Our previous works~\cite{bergmann_transportation_2026,bergmann_magbotsim_2026} are closest to this paper. We presented the 6D-Platform MagBot~\cite{bergmann_transportation_2026}, a reconfigurable 6-DoF kinematic mechanism that couples two movers. This MagBot has a platform with 6 DoF, thereby expanding the reachable workspace and payload of a single mover. In this paper, we extend this line of research by presenting a new MagBot with grasping capabilities that is compatible with the 6D-Platform MagBot.\\
\textbf{Reconfigurable Robots.} 
In combination with the MagLev system, our \emph{Gripper MagBot} is a reconfigurable robot, as it has a docking mechanism that enables autonomous pick-up and drop-off. However, there are two major differences compared to other reconfigurable robots, such as FreeBOT~\cite{liang_freebot_2020}, M-TRAN III~\cite{kurokawa_distributed_2008}, PuzzleBot~\cite{yi_puzzlebots_2021}, Swarm-Bot~\cite{mondada_swarm-bot_2004}, Slimebot~\cite{shimizu_amoeboid_2009}, or 3D M-Blocks~\cite{romanishin_3d_2015}. Firstly, our \emph{Gripper MagBot} itself contains no electronics, such as communication modules, processors, batteries, or actuators. This is not required, since we use the movers as actuators together with all capabilities and information provided by the MagLev system, e.g. the positions and dynamics of all movers are globally accessible. Secondly, the reconfigurability of the \emph{Gripper MagBot} refers to the reconfiguration of a machine and not to a change in shape.\\
\textbf{Mobile Manipulators.} Our \emph{Gripper MagBot} is a mobile manipulator based on a MagLev system. We consider aerial~\cite{kamel_design_2016,keemink_mechanical_2012,zhao_design_2023} and wheeled~\cite{schwarz_nimbro_2017,fu_mobile_2025,kemp_design_2022} mobile manipulators to be closest to our work. Typical applications for aerial manipulators include assembly and maintenance tasks or package delivery, while wheeled mobile manipulators are used in disaster response, exploration, e.g. in hazardous environments, warehouse logistics, or service robotics. In contrast, our MagBot is designed to combine in-machine transportation and manipulation, while using a MagLev system as a base platform.\\
\section{DESIGN DECISIONS}
\begin{figure}
	\centering
    \includegraphics[width=0.95\linewidth]{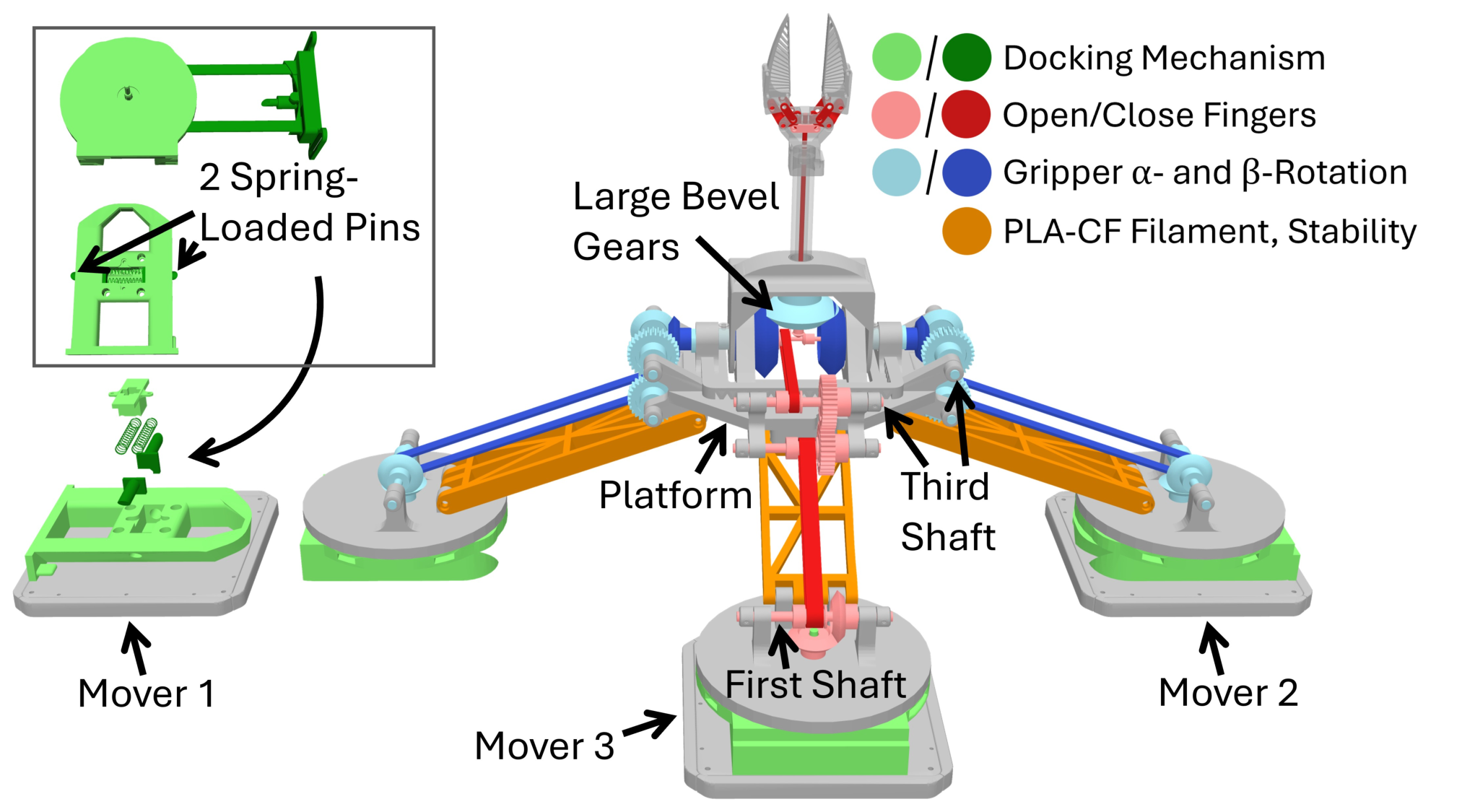}
        \caption{Visualization of mechanical components of the \emph{Gripper MagBot} required for the docking mechanism, the gripper's $\alpha$- and $\beta$-rotations, as well as the opening and closing of the fingers.}
    \label{fig_mechanics}
\end{figure}

\noindent
We designed the \emph{Gripper MagBot} with respect to the same conditions as the 6D-Platform MagBot~\cite{bergmann_transportation_2026}:
\begin{itemize}
    \item Small footprint (single-track mode)
    \item Self-sufficiency (no batteries, no actuation)
    \item Reconfigurability (pick-up/drop-off docking station)
    \item Compatibility (with all aforementioned MagLev systems and docking mechanism of 6D-Platform MagBot)
    \item Affordability (low-cost, lightweight, 3D-printed)
    \item Robustness (high dynamics, durability, repairability)
\end{itemize}
The primary components of the \emph{Gripper MagBot} are 3D-printed using PETG filament and $20\%$ infill, except for the connections depicted in orange in Fig.~\ref{fig_mechanics}. We used PLA-CF filament for these components to stabilize them and prevent torsional deformation, especially when the movers are rotating. The gripper fingers are 3D-printed from TPU filament to provide compliance. Since the movers have a limited payload capacity, all bevel and spur gears are made of plastic. Therefore, the \emph{Gripper MagBot} weighs $1.775\,$kg. In addition, the MagBot costs about $400\,$USD, making it low-cost. In this paper, $\alpha$, $\beta$, and $\gamma$ refer to the rotations about the x-, y-, and z-axes, respectively, as depicted in Fig.~\ref{fig_magbot_dofs_modes}.
\subsection{6-DoF Manipulator with 1-DoF Gripper}
\noindent
Similar to the 6D-Platform MagBot, the self-sufficiency of the \emph{Gripper MagBot} is achieved by using the MagLev movers as actuators, i.e. the DoFs of the gripper are mechanically coupled to the x-, y-, and $\gamma$-axes of the three movers. The $\gamma$-axes of the movers are used to control the $\alpha$- and $\beta$-rotations of the gripper, as well as the fingers, while the remaining DoFs of the gripper can be actuated using the x- and y-axes of the movers. 
More specifically, the MagBot moves in x- and y-directions by adjusting the x- and y-positions of the movers. To perform a $\gamma$-rotation of the gripper, the movers must be repositioned by moving them along a circle, again using their x- and y-axes, while simultaneously utilizing the movers' $\gamma$-rotations to maintain the desired $\alpha$- and $\beta$-rotations, as well as the desired finger opening. The gripper's z-position depends on the gripper's $\alpha$-rotation and the z-position of the platform (see Fig.~\ref{fig_mechanics}). The latter is adjusted by changing the distances between the movers, i.e. moving them in x- and y-directions. Therefore, the movers are mechanically coupled to the platform via the connections depicted in orange in Fig.~\ref{fig_mechanics}, thus forming a truncated triangular pyramid. These connections are attached to the platform and mover components via revolute joints to allow for the required movements when adjusting the distances between the movers. 
The components used to couple the $\gamma$-rotations of the movers with the $\alpha$- and $\beta$-rotations of the gripper are colored in blue in Fig.~\ref{fig_mechanics}, while the components that couple the mover's rotation with the gripper fingers are shown in red. The general mechanism for transmitting the $\gamma$-rotations of the movers to the platform is similar for all three movers: Similar to the 6D-Platform MagBot, we use ball bearings between the components colored in light green and gray on top of the movers. A bevel gear (light blue/light red) is mounted on a shaft (light green) that belongs to the light green components and sticks through the gray parts. A second bevel gear transfers the movers' $\gamma$-rotations to the lower shaft while rotating its direction by $90^\circ$. The rotation is further transmitted using a toothed belt that drives a second shaft with a spur gear that is connected to a second spur gear to actuate a third shaft. From now on, the mechanical components for the fingers and the gripper's rotations ($\alpha$ and $\beta$) are different. For the latter, a bevel gear is mounted on the third shaft opposite to the spur gear (light blue). This bevel gear is again connected to a second bevel gear (dark blue) to redirect the rotational direction by $90^\circ$. The second bevel gear drives a shaft, at the end of which a larger bevel gear (dark blue) is mounted. Starting from mover 1 and mover 2, the rotations are now transmitted to the large bevel gears depicted in dark blue in the center of the platform, which are connected via a third bevel gear shown in light blue. The latter is connected to the gripper. We chose this mechanism with the three large bevel gears to have enough space in the center to transmit the rotational motion of the third mover to open and close the fingers. If both movers rotate in opposite directions, the dark blue bevel gears rotate in similar directions, thereby controlling the $\alpha$-rotation of the gripper. If, on the other hand, both movers rotate in the same direction, the dark blue bevel gears rotate in opposite directions, thereby controlling the $\beta$-rotation of the gripper. If only one of the two movers rotates, the gripper rotates simultaneously in $\alpha$ and $\beta$, i.e. simultaneous $\gamma$-rotations of mover 1 and mover 2 are required to move the gripper solely in $\alpha$ or $\beta$. To open and close the fingers, a second toothed belt (dark red) transmits the rotational motion of the third shaft to a smaller shaft between the large bevel gears, which drives a very small bevel gear (light red). The latter is connected to another small bevel gear to redirect the rotational direction by $90^\circ$ in order to drive the dark red shaft that is connected to the fingers of the gripper. This shaft has a threaded end which moves the component visualized in light red up and down, thereby opening and closing the fingers of the gripper. 
\subsection{Compatibility and Docking Mechanism}
\label{sec_docking_mech}
\noindent
Since our goal is to equip MagLev systems with various manipulation capabilities that the movers can intelligently obtain by using MagBots, our objective was to design the \emph{Gripper MagBot} such that it is compatible with the 6D-Platform MagBot. More specifically, we used the same docking mechanism and docking stations with one minor modification: We added a second spring-loaded pin (see Fig.~\ref{fig_mechanics}, top left) to ensure that movers carrying a MagBot can switch roles. In the version of the 6D-Platform MagBot, each mover can only be used for its dedicated role, i.e. controlling the $\alpha$- or $\beta$-rotation, since the component of the docking mechanism that is attached to the mover is not rotationally symmetric. By adding the second pin, we enable role switching between the movers by achieving rotational symmetry of this component. Similar to the 6D-Platform MagBot, the \emph{Gripper MagBot} is compatible with all above-named MagLev systems without major changes, i.e. only the position of the screw holes in the component attached to the mover must be aligned with the movers' design.
\subsection{Single-Track Mode}
\label{sec_singletrackmode}
\noindent
The \emph{Gripper MagBot} can move in two configurations, the default mode or the single-track mode, depending on the required stability and workspace footprint, as shown in Fig.~\ref{fig_magbot_dofs_modes} (top right). In the single-track mode, the third mover moves below the platform, thereby reducing the footprint of the \emph{Gripper MagBot} in the y-direction. Note that we use the term single-track mode for all configurations in which the third mover is below the platform, even though the movers are not necessarily arranged in a straight line. The MagBot is generally more stable in default mode than in single-track mode for two reasons. Firstly, the third mover stabilizes the MagBot in the y-direction, and secondly, we can use different low-level control parameters (see Sec.~\ref{sec_control_params}). The single-track mode is primarily intended to allow the MagBot to move along a narrow one- or two-tile-wide path while allowing other movers to pass in the opposite direction. It is not primarily intended for manipulation, except when a small y-direction footprint is required.

\begin{figure}
	\centering
    \includegraphics[width=\linewidth]{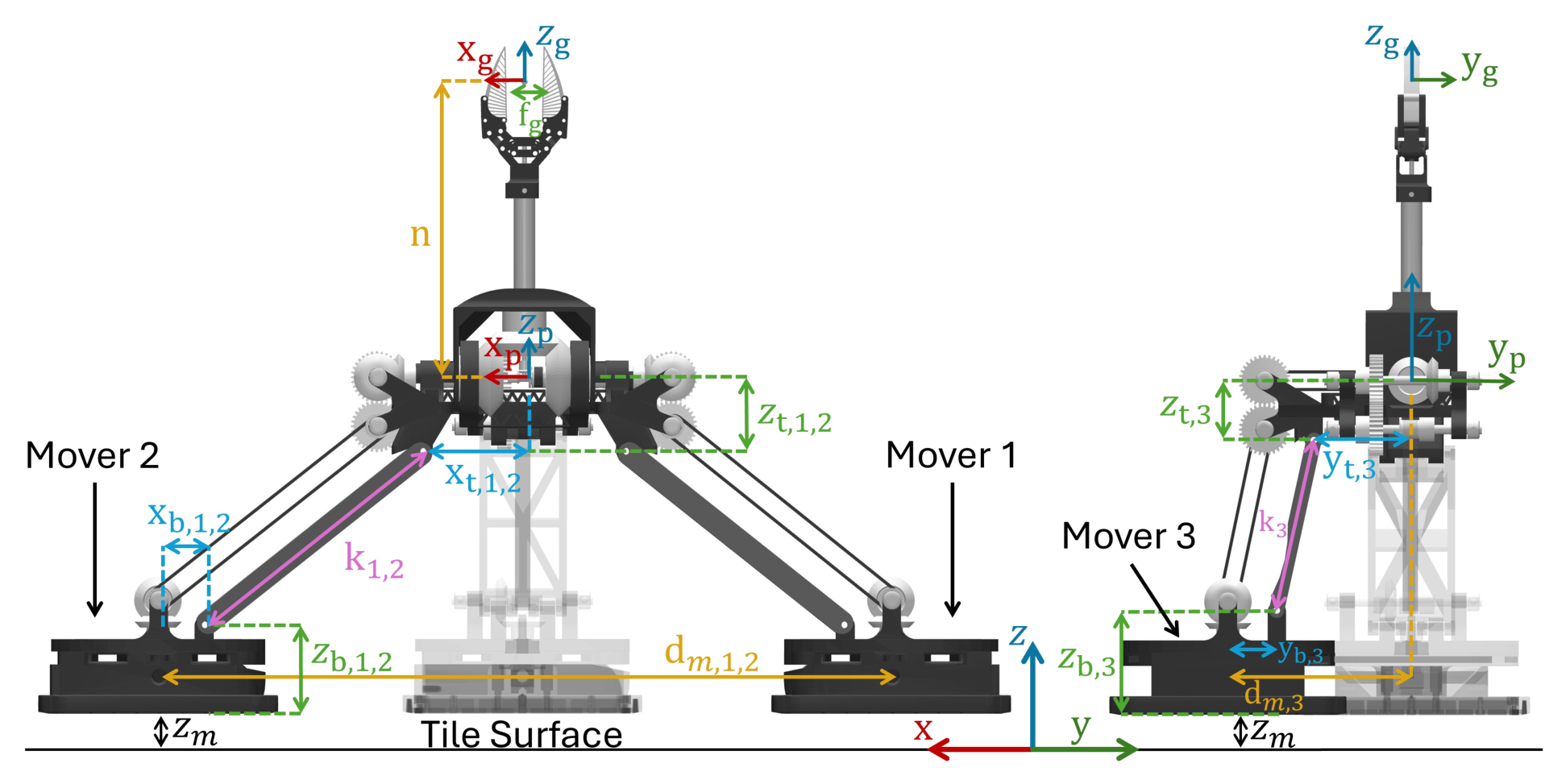}
        \caption{Coordinate frames and variables used for the inverse kinematics controller of the \emph{Gripper MagBot}.}
    \label{fig_magbot_vars}
\end{figure}
\section{INVERSE KINEMATICS CONTROLLER}
\label{sec_ikc}
\noindent
We developed an inverse kinematics controller that receives the desired end-effector pose (position, orientation, and distance between the gripper fingers) as input and outputs the required x- and y-positions, and $\gamma$-rotation of each mover. Since the movers' $\alpha$- and $\beta$-rotations are not used, they are set to $0^{\circ}$. Throughout this paper, subscripts $m$, $p$, and $g$ denote the desired positions of the movers, platform, and gripper, respectively. Moreover, subscripts $1,2,$ and $3$ are used to identify the three movers. Firstly, we calculate the desired position of the platform, which depends on the desired gripper $(x,y,z)$-position, the movers' flight altitude $z_m$, and the contribution of the gripper's $\alpha$-rotation:
\begin{equation}
    x_p = x_g + n\cdot\lvert\sin{(\alpha_g)}\rvert\cdot\sin{(\gamma_g)}
\end{equation}
\begin{equation}
    y_p = y_g - n\cdot\lvert\sin{(\alpha_g)}\rvert\cdot\cos{(\gamma_g)}
\end{equation}
\begin{equation}
    z_p = z_g - n\cdot\cos{(\alpha_g)} - z_m 
\end{equation}
\noindent
Secondly, we calculate the required distance between the centers of mover 1 and mover 2 $d_{m,1,2}$ to achieve the platform's z-position: 
\begin{equation}
    d_{m,1,2} = 2\sqrt{k_{1,2}^2-(z_p - z_{b,1,2} - z_{t,1,2})^2}+x_{b,1,2} + x_{t,1,2}
\end{equation}
where $k_{1,2},z_{b,1,2},z_{t,1,2},x_{b,1,2},$ and $x_{t,1,2}$ are known in advance from the CAD model (see Fig.~\ref{fig_magbot_vars}). In addition, the distance between the center of mover 3 and the platform must be adjusted accordingly:
\begin{equation}
    \label{eq_dm3}
     d_{m,3} = y_{b,3} + y_{t,3} \pm\sqrt{k_3^2 - (z_p - z_{b,3} - z_{t,3})^2}
\end{equation}
where $k_3,z_{b,3},z_{t,3},x_{b,3},$ and $x_{t,3}$ are again known from the CAD model (see Fig.~\ref{fig_magbot_vars}). The sign in front of the square root in Eq.~\ref{eq_dm3} depends on whether the \emph{Gripper MagBot} is being used in single-track mode, i.e. negative for single-track mode. Using these results, we can calculate the required x- and y-positions, as well as the $\gamma$-rotations of each mover, denoting the gear ratios as $s_{m,1},s_{m,2},$ and $s_{m,3}$.\\Mover 1:

\begin{equation}
    x_{m,1} = x_p - \frac{d_{m,1,2}}{2}\cdot\cos{(\gamma_g)}
\end{equation}
\begin{equation}
    y_{m,1} = y_p - \frac{d_{m,1,2}}{2}\cdot\sin{(\gamma_g)}
\end{equation}
\begin{equation}
    \label{eq_gamma_m1}
    \gamma_{m,1} = -s_{m,1}\cdot(\alpha_g + \beta_g) + \gamma_g
\end{equation}
Mover 2:
\begin{equation}
    x_{m,2} = x_p + \frac{d_{m,1,2}}{2}\cdot\cos{(\gamma_g)}
\end{equation}
\begin{equation}
    y_{m,2} = y_p + \frac{d_{m,1,2}}{2}\cdot\sin{(\gamma_g)}
\end{equation}
\begin{equation}
    \label{eq_gamma_m2}
    \gamma_{m,2} = s_{m,2}\cdot(\alpha_g - \beta_g) + \gamma_g
\end{equation}
Mover 3:
\begin{equation}
    x_{m,3} = x_p + d_{m,3}\cdot\sin{(\gamma_g)}
\end{equation}
\begin{equation}
    y_{m,3} = y_p - d_{m,3}\cdot\cos{(\gamma_g)}
\end{equation}
\begin{equation}
    \label{eq_gamma_m3}
    \gamma_{m,3} = s_{m,3}\cdot f_g + \alpha_g + \gamma_g
\end{equation}
Adding $\gamma_g$ to Eqs.~\ref{eq_gamma_m1} and \ref{eq_gamma_m2}, as well as adding $\alpha_g$ and $\gamma_g$ to Eq.~\ref{eq_gamma_m3}, compensates for rotations and finger movements caused by the gripper's $\alpha$- and $\gamma$-rotation. Our inverse kinematics controller can be used in combination with any trajectory planning approach that guarantees compliance with minimum and maximum mover distances, as well as minimum gripper positions depending on the current platform's z-height, to prevent damage or collisions with the tiles.
\section{PERFORMANCE METRICS}
\label{sec_performance_metrics}
\begin{table}
    \centering
    \scriptsize
    \caption{PERFORMANCE METRICS}
    \begin{tabular}{|c|c||c|c|}
        \hline
        \multicolumn{2}{|c||}{\textbf{Workspace}} &  \textbf{Max. Velocity} & $2\,$m/s \\
        \hline
        x & Limited by tile setup & \textbf{Max. Acceleration} & $10\,$m/s$^2$\\
        \hline
        y & Limited by tile setup & \textbf{Max. Payload} & $0.2\,$kg\\
        \hline
        z & $0\,$mm - $413.37\,$mm & \textbf{Weight} & $1.775\,$kg\\
        & (max. with $\alpha=0$) &  (without movers) & \\
        \hline
        $\alpha$ & $-180^\circ$ - $0^\circ$ & \textbf{Weight to Payload Ratio} & $8.875$\\
        \cline{1-2}
        $\beta$ & unlimited & (without movers) & \\
        \hline
        $\gamma$ & unlimited \\
        \cline{1-2}
        fingers & $0\,$mm - $17\,$mm\\
        \cline{1-2} 
    \end{tabular}
    \label{tab_performance_metrics}
\end{table}
\noindent
We provide different performance metrics for the \emph{Gripper MagBot}, such as the workspace, maximum dynamics, and information about payloads, as listed in Tab.~\ref{tab_performance_metrics}. In both operating modes, the \emph{Gripper MagBot} can move at a maximum velocity of $2\,$m/s and a maximum acceleration of $10\,$m/s$^2$. In x- and y-directions, the workspace depends on the current configuration of the MagLev system, as the tiles can be assembled in almost any layout. The workspace in z depends on both the position of the platform and the desired $\alpha$-rotation of the gripper. We provide the maximum possible workspace in z when using both, i.e. the maximum is obtained when $\alpha=0$ (pointing upward) and the platform is at its maximum possible z-position ($220.37\,$mm). Mechanically, the gripper's workspace is unlimited in $\beta$ and $\gamma$. However, the workspace in $\gamma$ can be dependent on the MagLev system, as all three movers must compensate using their $\gamma$-rotations to maintain the desired $\alpha$- and $\beta$-rotations of the gripper, as well as the desired distance between the gripper fingers (see Eqs.~\ref{eq_gamma_m1},~\ref{eq_gamma_m2},~\ref{eq_gamma_m3}). Thus, a full $360^\circ$ $\gamma$-rotation of the gripper can only be achieved if the MagLev system supports $360^\circ$ rotations of the movers at every position on a tile. The XPlanar system in our lab with APS4322-0000-0000 tiles currently only supports a $360^\circ$ rotation of the movers at specific positions. At all other positions, the movers can only achieve a $\gamma$-rotation of $\pm10^\circ$. However, as this is a limitation of the MagLev system with a specific tile type and not a constraint of the MagBot, we specify the workspace in $\gamma$ as unlimited.
\begin{figure}
	\centering
    \includegraphics[width=\linewidth]{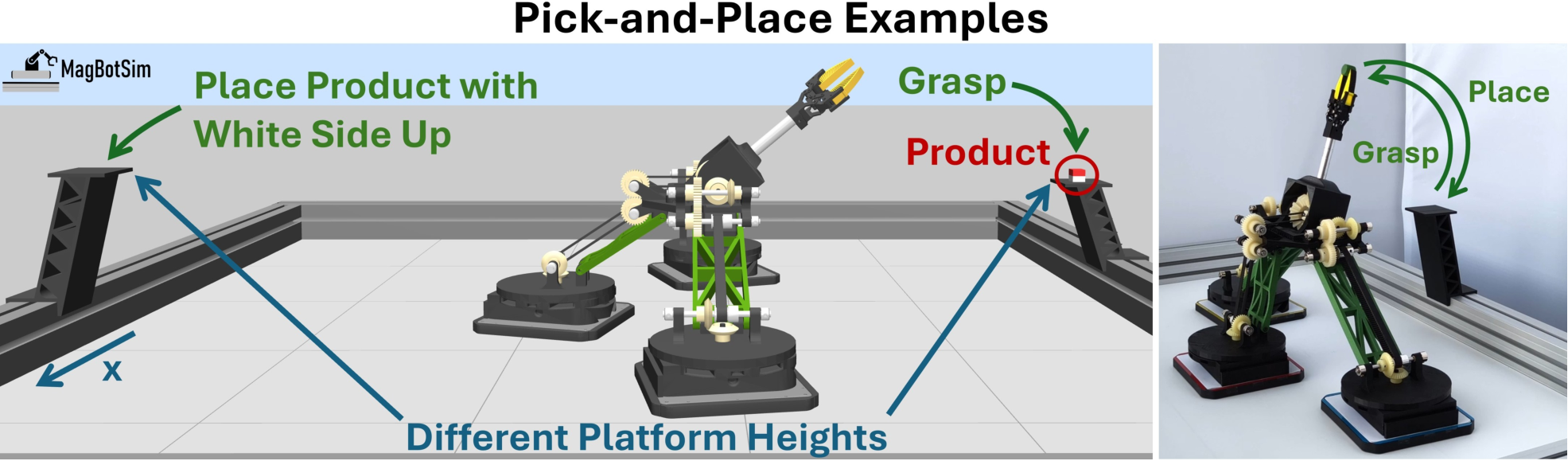}
        \caption{Pick-and-place tasks in simulation (left) and with the real MagBot (right). The tasks differ slightly due to limitations of the MagLev system.}
    \label{fig_pick_and_place_examples}
\end{figure}
\section{SIMULATION}
\noindent
We created a MuJoCo~\cite{todorov_mujoco_2012} model of the \emph{Gripper MagBot} that, along with the inverse kinematic controller, is integrated into the MagBotSim~\cite{bergmann_magbotsim_2026} library. Since MagBotSim is open-source, available via pip, and already includes the MuJoCo model of the 6D-Platform MagBot~\cite{bergmann_transportation_2026}, it facilitates the development of motion planning and manipulation approaches specifically designed for MagBots. Code and documentation for the \emph{Gripper MagBot} in simulation are available at \url{https://ubi-coro.github.io/MagBotSim/magbots.html}.\\
We showcase a pick-and-place example in simulation (see Fig.~\ref{fig_pick_and_place_examples}, left). The task is to pick up the cube (product), rotate it so that the white side is facing upwards, and place it onto the left platform. Since both platforms have different heights and are offset along the x-axis, this task requires all DoFs of the \emph{Gripper MagBot}. We measure the simulated task completion time, calculated as the number of control cycles multiplied by the cycle time. One run took $19.661\,$s, corresponding to a throughput of approximately $183$ products per hour. Further simulation experiments remain for future work.

\section{EXPERIMENTS}
\noindent
All experiments were conducted using a Beckhoff XPlanar MagLev system equipped with APM4330-0000-0000 movers and tiles of type APS4322-0000-0000 that are arranged in a 4x3 grid. We implemented our inverse kinematics controller using a TwinCAT 3 PLC. The following values (in mm) for the variables introduced in Sec.~\ref{sec_ikc} are measured using the CAD model: $n=193.0$, $z_{b,1,2}=58.11$, $z_{t,1,2}=48.4$, $x_{b,1,2}=30.0$, $x_{t,1,2}=66.0$, $z_{b,3}=68.01$, $z_{t,3}=38.5$, $x_{b,3}=32.0$, $x_{t,3}=64.0$, $k_{1,2}=182.0$, and $k_3=113.86$. To estimate the gear ratios $s_{m,1}$ and $s_{m,2}$, we rotated movers 1 and 2 from 0 to $40^\circ$ and measured the corresponding $\alpha$- and $\beta$-rotations of the gripper using a VICON motion capture system (MX T20 cameras, Nexus 2.6.1). Since $\beta=0^\circ$ during the whole movement, we set $s_{m,1}=s_{m,2}$. Using linear regression, we obtained $s_{m,1}=s_{m,2}=0.9969$. We estimated $s_{m,3}$ by rotating mover 3 in $360^\circ$ steps from $0^\circ$ to $1800^\circ$ and measured the distance between the gripper fingers at all multiples of $360^\circ$, again using the VICON system. Using linear regression, we obtained $s_{m,3}=-182.4376$. Unless stated otherwise, the desired mover flight altitude is set to $z_m=1\,$mm, since the movers are most stable at this z-position. Moreover, we use the default XPlanar low-level parameters (parameter set 0) for all three movers when the \emph{Gripper MagBot} is in default mode and we switch to parameter set 1 for the $\alpha$- and $\beta$-axes of the movers in single-track mode (see Sec.~\ref{sec_control_params} and Tab.~\ref{tab_control_params}).
\begin{table}
    \centering
    \scriptsize
    \caption{GRIPPER POSITIONING ACCURACY AND REPEATABILITY}
    \begin{tabular}{|c|c|c|}
        \hline
        \multicolumn{3}{|c|}{\textbf{Single-Axis Movements}}\\
        \hline
        \textbf{Axis} & \textbf{Default Mode} & \textbf{Single-Track Mode} \\
         (Pos., Repetitions, Min, Max) & & \\
        \hline
        \textbf{x} (y-10, 10, $335.0$, $625.0$) & $0.427\pm0.379\,$mm & $0.46\pm0.409\,$mm \\
        \hline
        \textbf{y} (x-10, 10, $248.0$, $641.0$) & $0.792\pm0.568\,$mm & $1.266\pm0.931\,$mm \\
        \hline
        \textbf{z} (xy-10, 10, $361.73$, $414.37$) & $0.441\pm0.322\,$mm & $0.357\pm0.338\,$mm \\
        
        \hline
        $\boldsymbol{\alpha}$ (y-2, 10, $-40$,$0$) & $2.141\pm1.625\,^\circ$ & --- \\
        \hline
        $\boldsymbol{\beta}$ (y-2, 3, $-30$,$30$) & $14.905\pm7.89\,^\circ$ & --- \\
        \hline
        $\boldsymbol{\gamma}$ (xy-10, 10, $-10$,$10$) & $0.084\pm0.064\,^\circ$ & $0.09\pm0.068\,^\circ$ \\
        \hline
        \textbf{fingers} (y-2, 3, $0$, $10$) & $0.544\pm0.508\,$mm & $0.975\pm1.173\,$mm \\
        \hline
        \multicolumn{3}{|c|}{\textbf{Multi-Axis Movements}}\\
        \hline
        \textbf{Dataset} & \textbf{Default Mode} & \textbf{Single-Track Mode} \\
        \hline
        x,y-Circle + Sine in z & $1.069\pm0.525\,$mm & $1.631\pm0.872\,$mm \\
        \hline 
        Helix & $2.621\pm2.265\,$mm & $4.569\pm4.717\,$mm \\
        \hline
    \end{tabular}
    \label{tab_accuracy_repeatability}
\end{table}
\subsection{Positioning Accuracy and Repeatability}
\begin{figure}
	\centering
    \includegraphics[width=\linewidth]{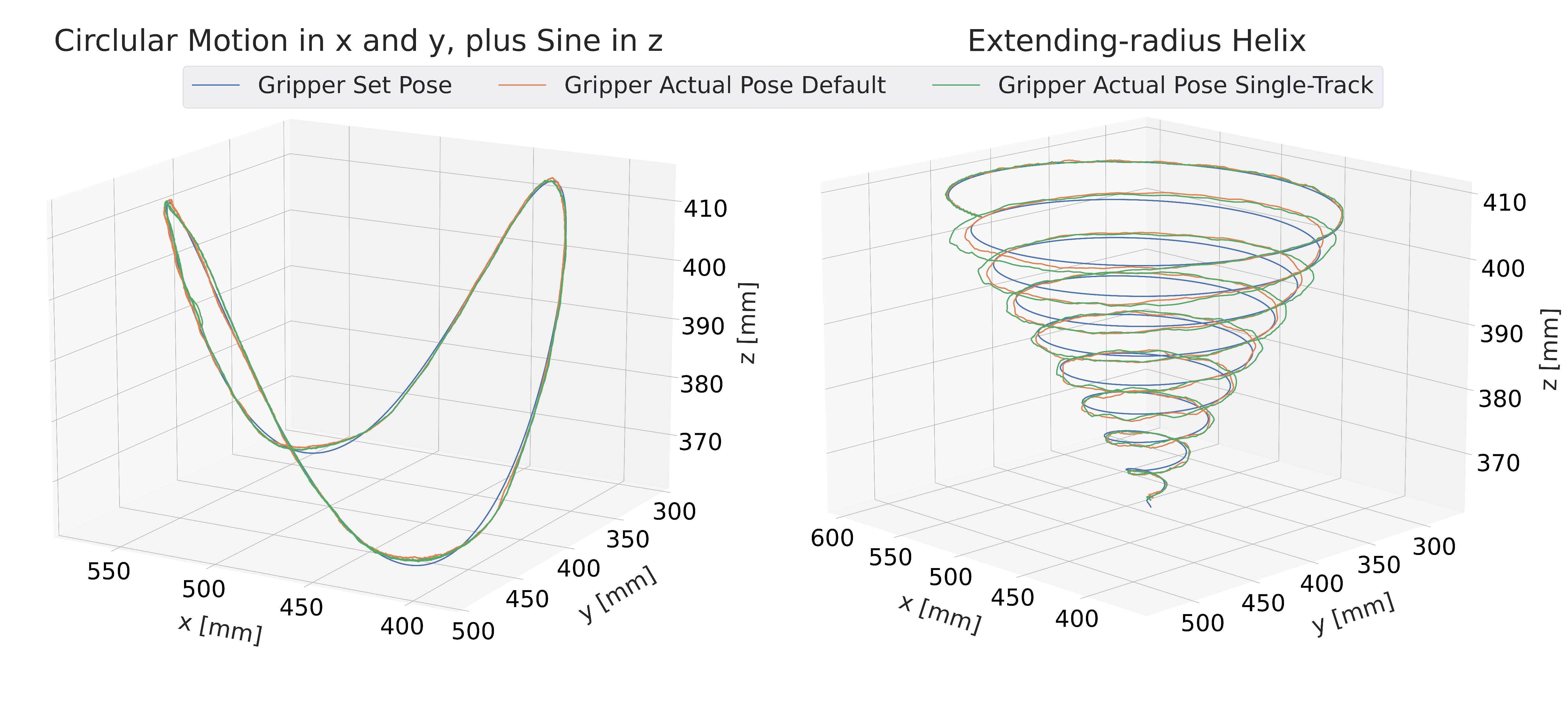}
        \caption{Simultaneous movements of the gripper in x, y, and z in default and single-track modes. The set positions are recorded in TwinCAT 3 and the actual positions are measured using a VICON motion capture system.}
    \label{plot_multiaxis_movements}
    \vspace{-0.2cm}
\end{figure}

\noindent
We measure the positioning accuracy (mean absolute error between the desired and actual positions) and repeatability (standard deviation) of each \emph{Gripper MagBot} axis in both operating modes, whenever possible, using the inverse kinematics controller. The VICON system tracked the gripper's actual pose (position and orientation), while the desired pose was recorded using TwinCAT 3. We moved each gripper axis from the minimum to the maximum position specified in Tab.~\ref{tab_accuracy_repeatability}, using the listed number of repetitions. The movements were repeated at multiple positions to account for position-dependent variations, summarized in Tab.~\ref{tab_accuracy_repeatability}, where, e.g. xy-10 denotes 10 different x- and y-positions. Due to the limitation of our XPlanar system, which only allows the $360^\circ$ $\gamma$-rotation of a mover at certain positions on a tile (see Sec.~\ref{sec_performance_metrics}), we only moved the gripper's $\gamma$-axis between $\pm10^\circ$ and repeated the movements for the $\alpha$- and $\beta$-axes, as well as the fingers, only at two different y-positions. For the same reason, measuring the $\alpha$-axis accuracy and repeatability in single-track mode was not possible, since at least one mover cannot perform a $360^\circ$ $\gamma$-rotation at the required position. During our experiments, we found another limitation of the XPlanar system, namely, the maximum torque that can be applied by the low-level controller depends on the current $\gamma$-rotation of the mover and the system's temperature, i.e. the higher the temperature, the less torque can be applied. The least torque can be applied at approximately $45^\circ$, $135^\circ$, $225^\circ$, $315^\circ$, and all corresponding multiples, which can lead to errors of the low-level controller, as setpoints may be unreachable. We would like to emphasize that this is a specific limitation of the XPlanar system and not of the \emph{Gripper MagBot}. Due to this limitation, we were not able to record data for $\beta$ in single-track mode. In addition, for $\alpha$, $\beta$, and the fingers, we chose the maximum possible min-max range that we could measure without errors of the low-level controller at the aforementioned $\gamma$-angles. In addition, to open and close the fingers, we had to increase the flight altitude of mover 1 and mover 2 to $3.5\,$mm, while the flight altitude of mover 3 remained unchanged, thereby removing the load from mover 3. Tab.~\ref{tab_accuracy_repeatability} shows that we can control the \emph{Gripper MagBot} with sub-millimeter/sub-degree accuracy in all axes, except for y in single-track mode, $\alpha$, and $\beta$. The larger errors in $\alpha$ are mainly caused by some mechanical play around the zero position of this axis. Similarly, the MagBot has some mechanical play in $\beta$, but the main reason for the large error is the MagLev system, as the low-level controller cannot generate the required torques (limitation described above) to rotate the $\beta$-axis precisely to its desired position. We tested the gripper's $\beta$-rotation by manually rotating the movers. This test revealed that the $\beta$-rotation works mechanically and that the error is therefore caused by the XPlanar system. Moreover, particularly for the y-axis and the fingers, the positioning accuracy and repeatability are better in default mode than in single-track mode. This can be explained by the greater stability of the MagBot in default mode due to the arrangement of the movers, which naturally stabilizes the entire kinematic, and the low-level control parameters. The latter have, among other differences, a larger $K_p$ value, causing the controller to react more strongly to positional deviations.
Using the VICON system, we also measured the positioning accuracy and repeatability of the \emph{Gripper MagBot} for simultaneous movements of the gripper in x, y, and z. More specifically, we analyzed two different trajectories, visualized in Fig.~\ref{plot_multiaxis_movements}, in both operating modes. The first trajectory is a circular motion in x and y superimposed with a sinusoidal motion in z, while the second trajectory is an extending-radius helix executed with high dynamics. The measured positioning accuracy and repeatability are summarized in Tab.~\ref{tab_accuracy_repeatability}. The results for the helix movement are worse than those for the first trajectory. This is not only a result of higher dynamics; the gripper also has some mechanical play in the y-direction, which leads to small movements in this direction (see Fig.~\ref{plot_multiaxis_movements}) that are more pronounced in this trajectory than in other movements due to its high dynamics. Similar to the single-axis experiments, the results in default mode are better than in single-track mode for both trajectories as a result of the different low-level control parameters and the mover arrangement. The difference is particularly large for the helix movement due to the high dynamics. 
\subsection{Control Parameters and Mode Switching}
\label{sec_control_params}
\begin{figure}
	\centering
    \includegraphics[width=0.79\linewidth]{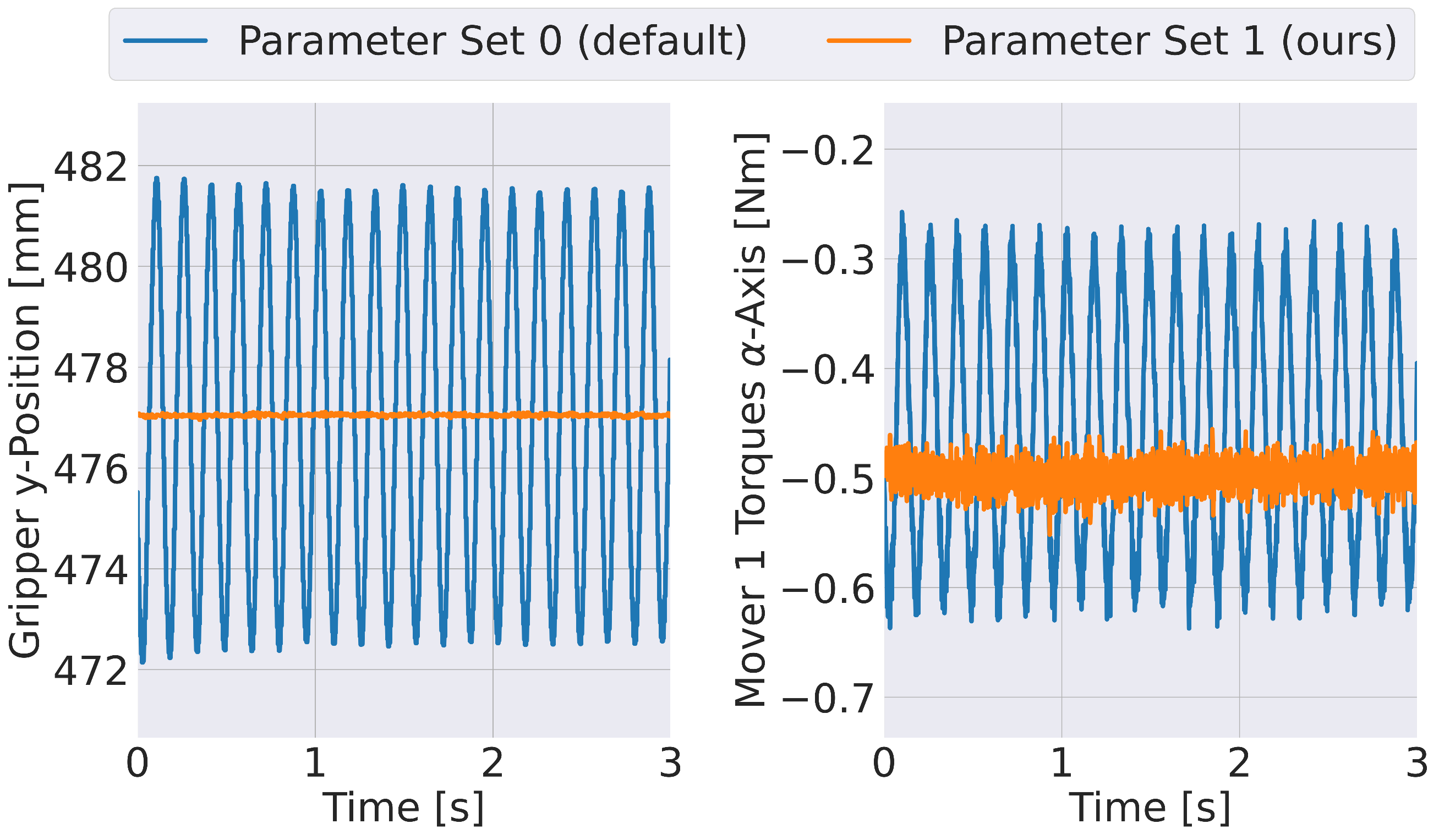}
        \caption{Gripper $y$-position (left) and mover torques of the $\alpha$-axis for low-level control parameter sets 0 and 1 without payload.}
    \label{plot_oscillation}
\end{figure}

\noindent
When switching to single-track mode, the default XPlanar low-level control parameters lead to significant oscillatory behavior. Similarly, we reported oscillations for the 6D-Platform MagBot in our previous work~\cite{bergmann_transportation_2026}. We found that parameter set 1 tuned for the 6D-Platform MagBot also clearly reduces the oscillations for the \emph{Gripper MagBot}. Thus, we use this parameter set in single-track mode for the $\alpha$- and $\beta$-axes of all movers, while the control parameters of the remaining mover axes are left at their default values. The specific values are listed in Tab.~\ref{tab_control_params} for both parameter set 0 and parameter set 1. It is not necessary to find an additional parameter set for the \emph{Gripper MagBot} with payload, as the MagBot remains stable in single-track mode with a payload when using parameter set 1. For the MagBot's default mode, the default control parameters do not cause oscillations, as the third mover stabilizes the MagBot. To show the importance of changing the control parameter sets when switching to single-track mode, we measured the gripper's y-position with the VICON system, as well as the torques the MagLev low-level controller applies to the $\alpha$-axis of mover 1, in single-track mode for both parameter sets. The results are shown in Fig.~\ref{plot_oscillation}. When using parameter set 0 in single-track mode, the oscillation is clearly visible in the gripper's y-position as well as in the $\alpha$-axis torques, while no oscillation is observable when using parameter set 1, i.e. the MagBot is stable. Since the oscillation is reflected in the movers’ torques, the low-level control parameters can be tuned without external feedback, such as a VICON system. This allows on-the-fly adjustment based solely on the movers’ torques, e.g. according to the current dynamics and configuration of the MagBot. An extensive investigation on how and when to adjust the control parameters is left for future work.
\begin{table}
    \centering
    \scriptsize
    \caption{CONTROL PARAMETER SETS FOR MOVER $\alpha$- AND $\beta$-AXES}
    \begin{tabular}{|c|c|c|c|}
        \hline
        Parameter Set & 0 (default) & 1 (ours) \\
        \hline 
        Purpose & MagBot Default Mode & MagBot Single-Track Mode \\
        \hline
        $K_p$ (Proportional Gain) & 35.0 & 25.0 \\
        \hline
        $T_n$ (Reset Time) & 0.03 & 0.12 \\
        \hline
        $T_v$ (Derivative Time) & 0.015 & 0.04 \\
        \hline
        $T_1$ (Derivative Delay) & 0.001 & 0.015\\
        \hline
    \end{tabular}
    \label{tab_control_params}
\end{table}
\subsection{Payloads and Mover Wrenches}
\begin{figure}
	\centering
    \includegraphics[width=\linewidth]{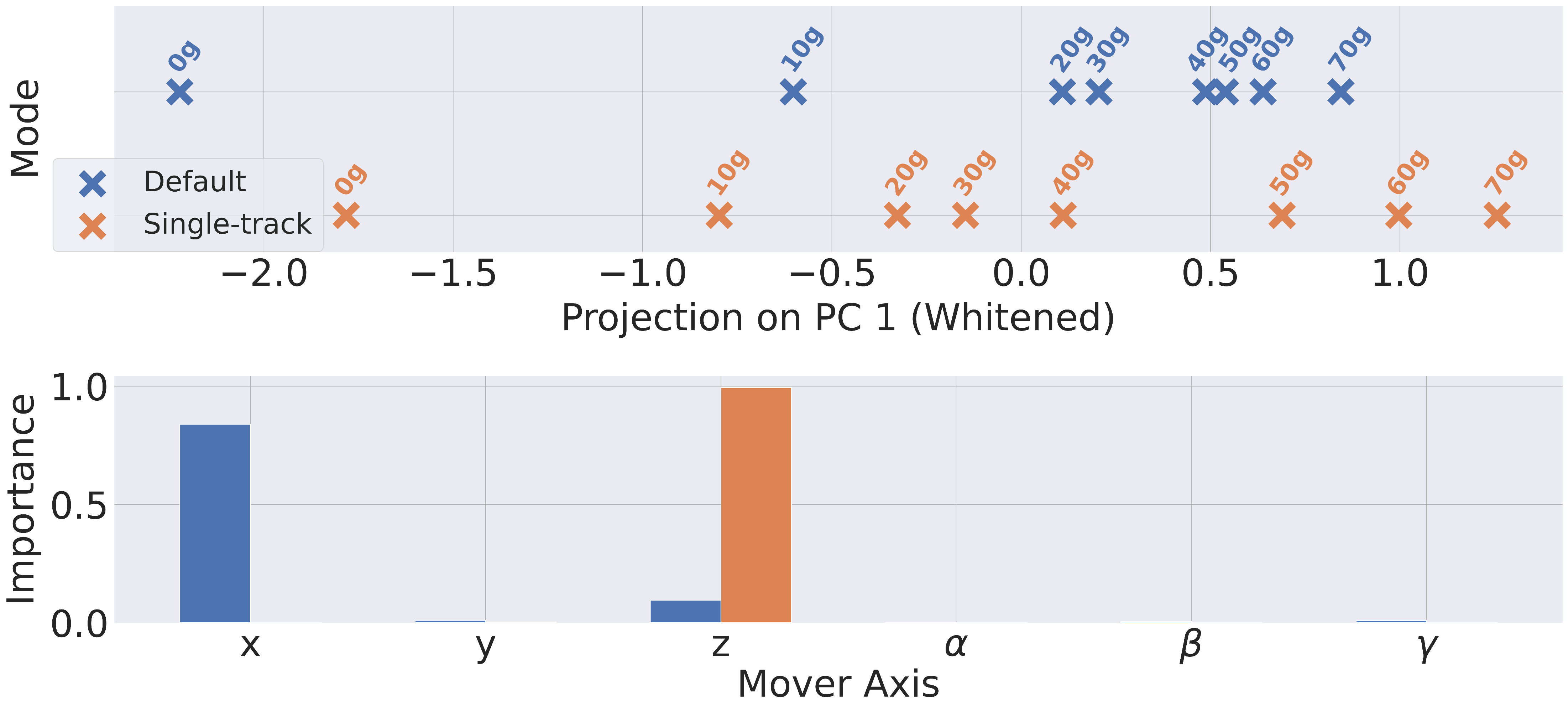}
        \caption{(Top) Projection of mean mover wrenches measured in default and single-track modes into a 1D space using PCA for dimensionality reduction (different PCA models per operating mode). (Bottom) Mover wrench component importance for each PCA model.}
    \label{plot_payloads}
    \vspace{-0.2cm}
\end{figure}

\noindent
We investigated whether the current payload of the \emph{Gripper MagBot} is related to the wrenches applied to the movers by the low-level MagLev controller in both modes of operation. To this end, we positioned payloads ranging from $0\,$g$\,$ to $\,70\,$g in increments of $10\,$g into the gripper and recorded the mover wrenches in default and single-track mode, i.e. we obtained 16 datasets (8 per mode). The movers did not move during the measurement. The gripper was pointing upward, as shown in Fig.~\ref{fig_magbot_dofs_modes}, and its set position was (480.0, 360.0, 361.73, 0, 0, 0) (in mm) for both default and single-track mode. The gripper was opened by rotating mover 3 in multiples of $360^\circ$ until the payload was held by the MagBot. For all datasets, we computed the mean mover wrenches $\mathcal{F}_{i,k} = \frac{1}{n}\sum^{n}_{l=1}\mathcal{F}_{i,k,l} \in \mathbb{R}^6$, where $i=1,...,16$ refers to the index of the dataset, $k=1,...3$ denotes the index of the mover, $n\in\mathbb{N}$ refers to the number of samples per dataset, and $\mathcal{F}_{i,k,l}$ denotes the l-th wrench of mover k in the i-th dataset. We reduced the dimensionality of the data to one dimension using principal component analysis~(PCA) on $\Delta_{\mathcal{F},d} = [\mathcal{F}_{1,1} - \mathcal{F}_{1,2}, ..., \mathcal{F}_{8,1} - \mathcal{F}_{8,2}]^T \in \mathbb{R}^{8x6}$
for the datasets collected in default mode ($i=1,...,8$) and on $\mathcal{F}_{s} = \left[\frac{1}{3}\sum^{3}_{k=1}\mathcal{F}_{9,k},...,\frac{1}{3}\sum^{3}_{k=1}\mathcal{F}_{16,k} \right]^T \in \mathbb{R}^{8x6}$
for the datasets recorded in single-track mode ($i=9,...,16$). The superscript $T$ indicates the transpose. It is not necessary to use a single PCA model for both operating modes to additionally estimate whether the MagBot is in default or single-track mode, as this information can always be derived from the current position of mover 3. We therefore use different PCA models for each operating mode. For both modes, the projected mover wrenches, shown in Fig.~\ref{plot_payloads} (top), are correctly arranged in ascending order. This result demonstrates that the wrenches of the movers contain a payload-dependent pattern. Since the desired gripper position must be maintained despite the additional weight, the MagLev low-level controller must counteract accordingly to keep the movers at their desired positions by applying additional forces and torques. Fig.~\ref{plot_payloads} (bottom) indicates that the PCA models mainly focus on the x- and z-elements of $\Delta_{\mathcal{F},d}$ in default mode and the z-component of $\mathcal{F}_{s}$ in single-track mode, i.e. these components are most affected by a changing payload. In single-track mode, the z-component is particularly crucial, as the third mover is positioned below the payload and the movers must keep their desired flight altitude. This also applies to the default mode, but in this mode the x-component is more important for the PCA model, as the desired distance in x-direction between mover 1 and mover 2 ($d_{m,1,2}$ in Fig.~\ref{fig_magbot_vars}) must also be maintained to keep the gripper at its desired z-position. Fig.~\ref{plot_payloads} (top) reveals that the default mode projections are often close to one another, whereas the single-track mode projections are clearly separated, since the third mover is positioned below the payload and is therefore particularly sensitive in the z-component. In default mode, however, all movers are offset from the payload in x- and y-directions, making it more difficult to distinguish between small weights. In summary, we showed that the wrenches of the movers contain payload-dependent patterns, enabling payload estimation solely from the movers' wrenches. However, during our experiments, the MagLev system heated up quickly. We therefore suggest investigating a possible temperature dependency or water cooling in future work.
\begin{figure}
	\centering
    \includegraphics[width=0.9\linewidth]{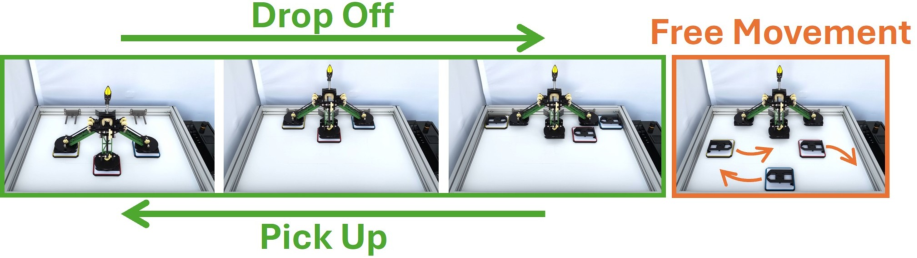}
        \caption{Visualization of the autonomous reconfigurability of our \emph{Gripper MagBot}. After drop-off, the movers can perform any movement (orange).}
    \label{fig_reconfigurability}
    \vspace{-0.3cm}
\end{figure}
\subsection{Reconfigurability}
\noindent
We used the same docking stations as for the 6D-Platform MagBot~\cite{bergmann_transportation_2026}, which are made from PETG with forks being reinforced with stainless steel plates. However, depending on the 3D printer, we had problems reproducing the reinforced forks. Therefore, we tested forks made from PPA-CF filament. Since PPA-CF is stronger than PETG, the PPA-CF forks are not reinforced, which simplifies the printing process. We found that it is not possible to use solely PPA-CF forks for all three docking stations, as they bend slightly more than the reinforced ones. Thus, the forks of the left and right docking stations in Fig.~\ref{fig_reconfigurability} must be reinforced with stainless steel plates, while the one in the center can be made of PPA-CF. For the following experiments, we used two reinforced docking stations and one made of PPA-CF. To evaluate the reconfigurability, we picked up and dropped off our \emph{Gripper MagBot} 15 times each. The movers switch positions between drop-off and pick-up during the free movement phase (see Fig.~\ref{fig_reconfigurability}) to demonstrate their interchangeability, which is achieved through our modification of the component screwed onto the mover (see Sec.~\ref{sec_docking_mech}). Thus, the \emph{Gripper MagBot} is picked up and dropped off five times in each mover configuration. We obtained $100\%$ success rates for both pick-up and drop-off, demonstrating that we can reliably use the docking mechanism for reconfiguration. 
\subsection{Pick-and-Place Example}
\noindent
In addition to the example in simulation, we showcase a pick-and-place example using the real \emph{Gripper MagBot}. As shown in Fig.~\ref{fig_pick_and_place_examples} (right), the MagBot picks up a cube (product) from a platform, moves around to account for a real trajectory with a grasped product, and places the cube back onto the same platform. This task requires using 4+1 DoFs (x,$\,$y,$\,$z,$\,\alpha$, and the fingers) of the MagBot. The task was completed in about $1.142\,$min, corresponding to roughly 52 products per hour. Due to the aforementioned limitation of the XPlanar system (see Sec.~\ref{sec_performance_metrics}), we again had to increase the flight altitude of movers 1 and 2 to $3.5\,$mm to open and close the fingers, while the flight altitude of mover 3 remained unchanged, thereby removing its load. Unlike the simulation, the real XPlanar system only allows the $360^\circ$ $\gamma$-rotation of the mover at specific positions on a tile (see Sec.~\ref{sec_performance_metrics}). Due to this limitation, we currently cannot perform a $180^\circ$ rotation of the MagBot on the real system. Therefore, the MagBot can currently pick up or place the cube at only one (x, y, z)-position on our MagLev system. Nevertheless, we demonstrated that the \emph{Gripper MagBot} can successfully grasp and place an object. Due to their reconfigurability, MagBots are particularly advantageous for high-mix, low-volume manufacturing scenarios, e.g. assembly tasks in the electronics/semiconductor or pharmaceutical industries.
\section{CONCLUSION AND FUTURE WORK}
\noindent
We introduced the \emph{Gripper MagBot}, a low-cost parallel 6-DoF manipulator with a 1-DoF gripper that is mechanically coupled to three MagLev movers. The MagBot has two operating modes, default mode and single-track mode, which differ only in terms of stability and workspace footprint. The \emph{Gripper MagBot} is compatible with the 6D-Platform MagBot~\cite{bergmann_transportation_2026}, as it is autonomously reconfigurable using the same docking mechanism, which makes it possible to equip MagLev systems with various manipulation capabilities that the movers can intelligently obtain by using MagBots. We found that the wrenches of the movers contain payload-dependent patterns, enabling payload estimation solely from the movers’ wrenches. However, this relation might also depend on the temperature of the MagLev system and could be further investigated in future work. Additionally, we demonstrated that the \emph{Gripper MagBot} achieves sub-millimeter/sub-degree positioning accuracy and repeatability for almost all axes in both operating modes using our inverse kinematics controller, except for the y-axis (single-track mode), $\alpha$, and $\beta$. The positioning accuracy in these axes, especially in $\beta$, should be improved in future work. In addition, future work should focus on enhancing the MagLev system, including its low-level controller, to reliably apply similar maximum torques at all possible $\gamma$-rotations of the movers. In conclusion, the \emph{Gripper MagBot} substantially extends the functional dexterity of a MagLev system by enabling grasping in Magnetic Robotics based solely on the MagLev system without requiring expensive handling equipment. 


\section*{ACKNOWLEDGMENT}
\noindent
Claude and ChatGPT have been used to accelerate debugging and to generate initial code to estimate the actual gripper pose from the VICON markers. The generated code has been adapted, reviewed, and tested by the authors. ChatGPT was also used to improve the writing and correct grammatical errors. All resulting changes were reviewed by the authors. Gemini and ChatGPT have been used to find related work. All the cited works have been reviewed by the authors.

\bibliographystyle{IEEEtran}
\bibliography{ReferencesGripperMagBot}

\begin{thebibliography}{10}
\providecommand{\url}[1]{#1}
\csname url@rmstyle\endcsname
\providecommand{\newblock}{\relax}
\providecommand{\bibinfo}[2]{#2}
\providecommand\BIBentrySTDinterwordspacing{\spaceskip=0pt\relax}
\providecommand\BIBentryALTinterwordstretchfactor{4}
\providecommand\BIBentryALTinterwordspacing{\spaceskip=\fontdimen2\font plus
\BIBentryALTinterwordstretchfactor\fontdimen3\font minus \fontdimen4\font\relax}
\providecommand\BIBforeignlanguage[2]{{%
\expandafter\ifx\csname l@#1\endcsname\relax
\typeout{** WARNING: IEEEtran.bst: No hyphenation pattern has been}%
\typeout{** loaded for the language `#1'. Using the pattern for}%
\typeout{** the default language instead.}%
\else
\language=\csname l@#1\endcsname
\fi
#2}}

\bibitem{lu_6d_2012}
X.~Lu and I.-u.-r. Usman, ``{6D} direct-drive technology for planar motion stages,'' \emph{CIRP Annals}, vol.~61, no.~1, pp. 359--362, 2012.

\bibitem{bergmann_transportation_2026}
L.~Bergmann, N.~Greis, and K.~Neumann, ``From {Transportation} to {Manipulation}: {Transforming} {Magnetic} {Levitation} to {Magnetic} {Robotics},'' \emph{arXiv:2603.01982}, 2026.

\bibitem{bergmann_magbotsim_2026}
L.~Bergmann, C.~Grothues, and K.~Neumann, ``{MagBotSim}: {Physics}-{Based} {Simulation} and {Reinforcement} {Learning} {Environments} for {Magnetic} {Robotics},'' in \emph{Proc. of the {Int}. {Conf}. on {Swarm} {Intelligence} ({ANTS})}, ser. Lecture {Notes} in {Computer} {Science} ({LNCS}), no. 16515.\hskip 1em plus 0.5em minus 0.4em\relax Springer, 2026, pp. 492--493.

\bibitem{isitman_trajectory_2025}
O.~Isitman, G.~Alcan, and V.~Kyrki, ``Trajectory {Planning} and {Control} for {Robotic} {Manipulation} of {Magnetic} {Capsules},'' \emph{IEEE Robotics and Automation Letters}, vol.~10, no.~5, pp. 4666--4673, 2025.

\bibitem{sallam_autonomous_2024}
M.~Sallam, M.~A. Shamseldin, and F.~Ficuciello, ``Autonomous navigation and control of magnetic microcarriers using potential field algorithm and adaptive non-linear {PID},'' \emph{Frontiers in Robotics and AI}, vol.~11, p. 1439427, 2024.

\bibitem{liu_on_2024}
Y.~Liu, Z.~Hou, and Q.~Fan, ``On a {Magnetically} {Driven} {Array} {System} with {Autonomous} {Motion} and {Object} {Delivery} for {Biomedical} {Microrobots},'' in \emph{Proc. of the {IEEE/RSJ} {Int.} {Conf.} on {Intelligent} {Robots} and {Systems} ({IROS})}, 2024, pp. 1357--1362.

\bibitem{hartmann_end--end_2025}
P.~Hartmann, J.~Stranghöner, and K.~Neumann, ``End-to-{End} {Low}-{Level} {Neural} {Control} of an {Industrial}-{Grade} {6D} {Magnetic} {Levitation} {System},'' \emph{arXiv:2509.01388}, 2025.

\bibitem{pierer_von_esch_sensitivity-based_2025}
M.~Pierer Von~Esch, E.~Nistler, A.~Völz, and K.~Graichen, ``Sensitivity-{Based} {Distributed} {NMPC}: {Experimental} {Results} for a {Levitating} {Planar} {Motion} {System},'' \emph{IEEE Transactions on Control Systems Technology}, vol.~33, no.~3, pp. 1110--1118, 2025.

\bibitem{janning_conflict-based_2025}
K.~Janning, A.~Housin, C.~Schulte, F.~Erkens, L.~Frenken, L.~Herbst, B.~Nießing, and R.~H. Schmitt, ``Conflict-based model predictive control for multi-agent path finding experimentally validated on a magnetic planar drive system,'' \emph{Frontiers in Control Engineering}, vol.~6, p. 1645918, 2025.

\bibitem{tistaert_multi-agent_2026}
B.~Tistaert, S.~Servaes, A.~Gonzalez-Garcia, I.~Ibrahim, L.~Callens, J.~Swevers, and W.~Decré, ``Multi-{Agent} {Motion} {Planning} on {Industrial} {Magnetic} {Levitation} {Platforms}: {A} {Hybrid} {ADMM}-{HOCBF} approach,'' \emph{arXiv:2603.19838}, 2026.

\bibitem{liang_freebot_2020}
G.~Liang, H.~Luo, M.~Li, H.~Qian, and T.~L. Lam, ``{FreeBOT}: {A} {Freeform} {Modular} {Self}-reconfigurable {Robot} with {Arbitrary} {Connection} {Point} - {Design} and {Implementation},'' in \emph{Proc. of the {IEEE}/{RSJ} {Int}. {Conf}. on {Intelligent} {Robots} and {Systems} ({IROS})}, 2020, pp. 6506--6513.

\bibitem{kurokawa_distributed_2008}
H.~Kurokawa, K.~Tomita, A.~Kamimura, S.~Kokaji, T.~Hasuo, and S.~Murata, ``\BIBforeignlanguage{en}{Distributed {Self}-{Reconfiguration} of {M}-{TRAN} {III} {Modular} {Robotic} {System}},'' \emph{\BIBforeignlanguage{en}{The International Journal of Robotics Research}}, vol.~27, no. 3-4, pp. 373--386, 2008.

\bibitem{yi_puzzlebots_2021}
S.~Yi, Z.~Temel, and K.~Sycara, ``{PuzzleBots}: {Physical} {Coupling} of {Robot} {Swarms},'' in \emph{Proc. of the {IEEE} {Int}. {Conf}. on {Robotics} and {Automation} ({ICRA})}, 2021, pp. 8742--8748.

\bibitem{mondada_swarm-bot_2004}
F.~Mondada, G.~C. Pettinaro, A.~Guignard, I.~W. Kwee, D.~Floreano, J.-L. Deneubourg, S.~Nolfi, L.~M. Gambardella, and M.~Dorigo, ``\BIBforeignlanguage{en}{Swarm-{Bot}: {A} {New} {Distributed} {Robotic} {Concept}},'' \emph{\BIBforeignlanguage{en}{Autonomous Robots}}, vol.~17, no. 2-3, pp. 193--221, 2004.

\bibitem{shimizu_amoeboid_2009}
M.~Shimizu and A.~Ishiguro, ``An amoeboid modular robot that exhibits real-time adaptive reconfiguration,'' in \emph{Proc. of the {IEEE}/{RSJ} {Int}. {Conf}. on {Intelligent} {Robots} and {Systems} ({IROS})}, 2009, pp. 1496--1501.

\bibitem{romanishin_3d_2015}
J.~W. Romanishin, K.~Gilpin, S.~Claici, and D.~Rus, ``{3D} {M}-{Blocks}: {Self}-reconfiguring robots capable of locomotion via pivoting in three dimensions,'' in \emph{Proc. of the {IEEE} {Int}. {Conf}. on {Robotics} and {Automation} ({ICRA})}, 2015, pp. 1925--1932.

\bibitem{kamel_design_2016}
M.~Kamel, K.~Alexis, and R.~Siegwart, ``Design and modeling of dexterous aerial manipulator,'' in \emph{Proc. of the {IEEE}/{RSJ} {Int}. {Conf}. on {Intelligent} {Robots} and {Systems} ({IROS})}, 2016, pp. 4870--4876.

\bibitem{keemink_mechanical_2012}
A.~Keemink, M.~Fumagalli, S.~Stramigioli, and R.~Carloni, ``Mechanical design of a manipulation system for unmanned aerial vehicles,'' in \emph{Proc. of the {IEEE} {Int}. {Conf}. on {Robotics} and {Automation} ({ICRA})}, 2012, pp. 3147--3152.

\bibitem{zhao_design_2023}
M.~Zhao, T.~Anzai, and T.~Nishio, ``Design, {Modeling}, and {Control} of a {Quadruped} {Robot} {SPIDAR}: {Spherically} {Vectorable} and {Distributed} {Rotors} {Assisted} {Air}-{Ground} {Quadruped} {Robot},'' \emph{IEEE Robotics and Automation Letters}, vol.~8, no.~7, pp. 3923--3930, 2023.

\bibitem{schwarz_nimbro_2017}
M.~Schwarz, T.~Rodehutskors, D.~Droeschel, M.~Beul, M.~Schreiber, N.~Araslanov, I.~Ivanov, C.~Lenz, J.~Razlaw, S.~Schüller, D.~Schwarz, A.~Topalidou‐Kyniazopoulou, and S.~Behnke, ``\BIBforeignlanguage{en}{{NimbRo} {Rescue}: {Solving} {Disaster}‐response {Tasks} with the {Mobile} {Manipulation} {Robot} {Momaro}},'' \emph{\BIBforeignlanguage{en}{Journal of Field Robotics}}, vol.~34, no.~2, pp. 400--425, 2017.

\bibitem{fu_mobile_2025}
Z.~Fu, T.~Z. Zhao, and C.~Finn, ``Mobile {ALOHA}: {Learning} {Bimanual} {Mobile} {Manipulation} using {Low}-{Cost} {Whole}-{Body} {Teleoperation},'' in \emph{Proc. of the {Conf}. on {Robot} {Learning}}, vol. 270, 2025, pp. 4066--4083.

\bibitem{kemp_design_2022}
C.~C. Kemp, A.~Edsinger, H.~M. Clever, and B.~Matulevich, ``The {Design} of {Stretch}: {A} {Compact}, {Lightweight} {Mobile} {Manipulator} for {Indoor} {Human} {Environments},'' in \emph{Proc. of the {Int}. {Conf}. on {Robotics} and {Automation} ({ICRA})}, 2022, pp. 3150--3157.

\bibitem{todorov_mujoco_2012}
E.~Todorov, T.~Erez, and Y.~Tassa, ``{MuJoCo}: {A} physics engine for model-based control,'' in \emph{Proc. of the {IEEE}/{RSJ} {Int}. {Conf}. on {Intelligent} {Robots} and {Systems} ({IROS})}, 2012, pp. 5026--5033.

\end{thebibliography}

\end{document}